\documentclass{article}
\usepackage{amssymb}

\usepackage[final]{corl_2025} 

\PassOptionsToPackage{numbers,sort&compress,square}{natbib}
\usepackage{multicol}
\usepackage{enumitem}  
\usepackage{float}
\usepackage{afterpage}
\usepackage{stfloats}
\usepackage{tabularx}
\usepackage{array}
\usepackage{makecell}
\usepackage{graphicx}
\usepackage{booktabs}
\usepackage{multirow}
\usepackage{dsfont}
\usepackage[dvipsnames]{xcolor}
\definecolor{mydarkblue}{rgb}{0,0.08,0.45}
\usepackage{xspace}
\usepackage{amsmath} 
\usepackage{amssymb}  
\usepackage{cite}
\usepackage{cuted}
\usepackage{caption}
\usepackage{bbm}
\usepackage{glossaries}
\usepackage{lipsum}

\glsdisablehyper

\usepackage{mdframed} 
\usepackage[T1]{fontenc} 
\usepackage{wrapfig}

\newcommand{\secref}[1]{Section~\ref{#1}}
\renewcommand{\eqref}[1]{Eqn~\ref{#1}}
\newcommand{\figref}[1]{Figure~\ref{#1}}

\newcommand{\tabref}[1]{Table~\ref{#1}}

\newcommand{\appendref}[1]{Appendix~\ref{#1}}

\newcommand{\ie}{\textrm{i.e.}}
\newcommand{\eg}{\textrm{e.g.}}

\newacronym{rl}{RL}{Reinforcement Learning}
\newacronym{drl}{DRL}{Deep Reinforcement Learning}
\newacronym{urma}{URMA}{Unified Robot Morphology Architecture}
\newacronym{gnn}{GNN}{Graph Neural Network}
\newacronym{ppo}{PPO}{Proximal Policy Optimization}
\newacronym{mpc}{MPC}{Model Predictive Control}
\newacronym{mlp}{MLP}{Multilayer Perceptron}
\newacronym{mdp}{MDP}{Markov Decision Process}
\newacronym[\glslongpluralkey={Degrees of Freedom}, \glsshortpluralkey={DoFs}]{dof}{DoF}{Degrees of Freedom}
\newacronym{mse}{MSE}{Mean Squared Error}
\newacronym{tsne}{t-SNE}{t-distributed Stochastic Neighbor Embedding}
\newacronym{pca}{PCA}{Principal Component Analysis}
\newacronym{umap}{UMAP}{Uniform Manifold Approximation and Projection}

\newcommand{\bbone}{\text{\usefont{U}{bbold}{m}{n}1}}
\MakeRobust{\bbone}

\newcommand{\curve}[1]{C#1}

\def\ourdataset{\textsc{GenBot-1K}\xspace}

\def\weblink{\urllink[pre = \bgroup\bf, post = \egroup]}

  {\begin{mdframed}[topline=false, bottomline=false, rightline=false,%
    linewidth=1pt,linecolor=black,%
    leftmargin=1cm,rightmargin=1cm,%
    innerleftmargin=10pt,innerrightmargin=10pt,%
    innertopmargin=0pt,innerbottommargin=0pt,%
    skipabove=\baselineskip,skipbelow=\baselineskip]%
   \fontfamily{ppl}\selectfont
  }%
  {\end{mdframed}}

\definecolor{DeepGreen}{rgb}{0.0, 0.5, 0.0}   
\definecolor{DeepBlue}{rgb}{0.0, 0.2, 0.6}    

\newcommand\blfootnote[1]{%
\begingroup 
\renewcommand\thefootnote{}\footnote{#1}%
\addtocounter{footnote}{-1}%
\endgroup 
}

\title{Towards Embodiment Scaling Laws in \\ Robot Locomotion \vspace{-6pt}
}

\author{
\textbf{Bo Ai}$^{1*}$ \quad \textbf{Liu Dai}$^{1*}$ \quad \textbf{Nico Bohlinger}$^{4*}$ \quad \textbf{Dichen Li}$^{1*}$ \quad \textbf{Tongzhou Mu}$^{1}$ \\ \quad \textbf{Zhanxin Wu}$^{3}$ \quad
\textbf{K. Fay}$^{1}$ \quad \textbf{Henrik I. Christensen}$^{1}$ \quad \textbf{Jan Peters}$^{4,5}$ \quad \textbf{Hao Su}$^{1, 2}$ \vspace{3pt} \\ 
$^{1}$University of California San Diego, USA \quad $^{2}$Hillbot Inc, USA \\
$^{3}$Cornell University, USA \quad $^{4}$Technical University of Darmstadt, Germany \\
$^{5}$German Research Center for AI (DFKI); Robotics Institute Germany; hessian.AI, Germany \vspace{2pt} \\
\href{https://embodiment-scaling-laws.github.io/}{\textbf{\textcolor{magenta}{https://embodiment-scaling-laws.github.io}}}
\vspace{-2pt} 
}

\begin{document}
\maketitle


\vspace{-28pt}
\begin{figure*}[h]
    \centering
    \includegraphics[width=1.0\textwidth]{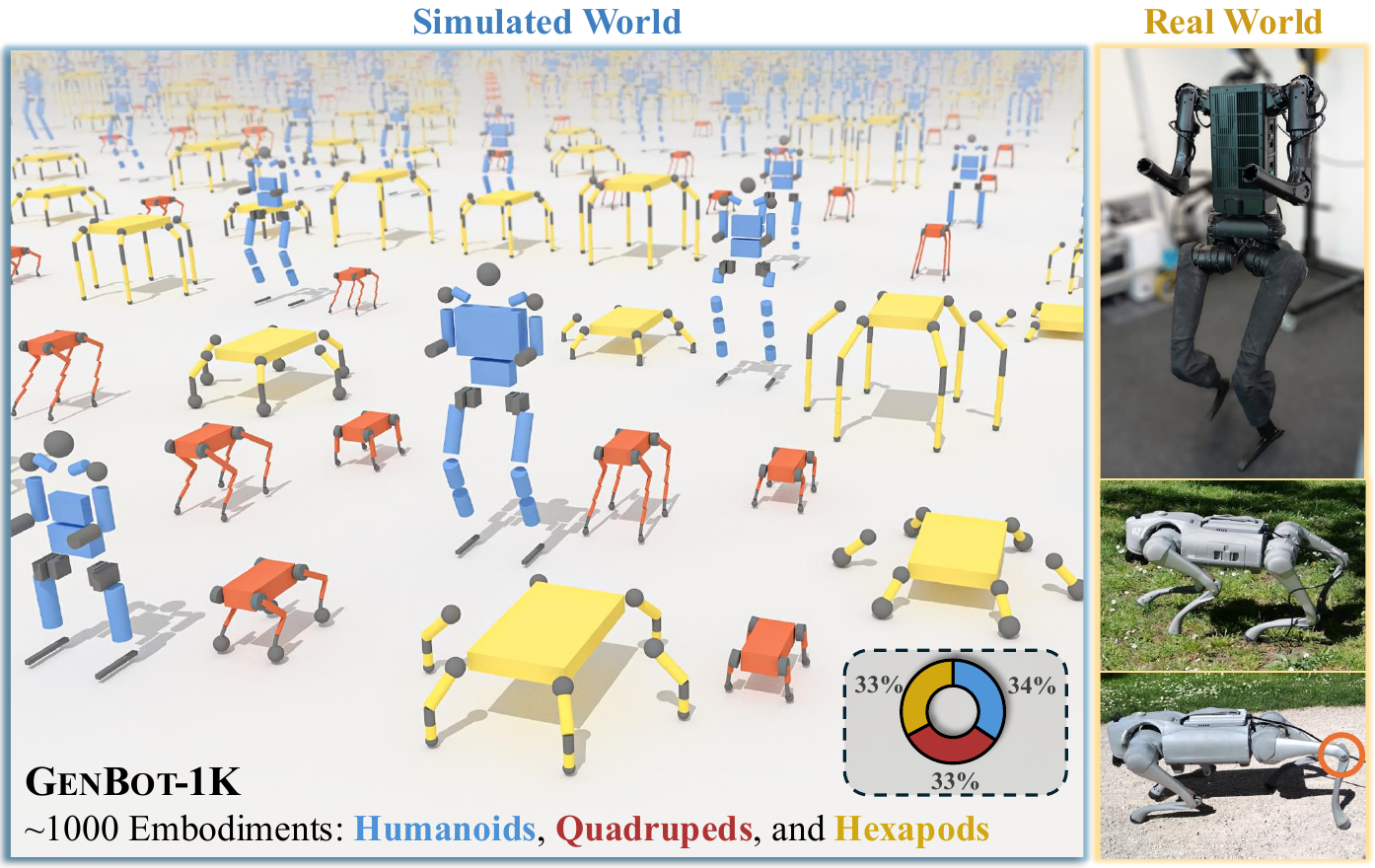} \vspace{-15pt}
    \caption{
        \textbf{One Policy, Two Worlds, Many Robots.} We study embodiment scaling laws by training a single policy on $\sim$1,000 procedurally generated ``blueprint'' embodiments in simulation. 
        Our policy zero-shot transfers to real-world embodiments, including modified joint constraints (circled in red).
    }
    \vspace{-16pt}
    \label{fig:teaser}
\end{figure*}


\begin{abstract} 
Cross-embodiment generalization underpins the vision of building generalist embodied agents for \emph{any} robot, yet its enabling factors remain poorly understood. We investigate \emph{embodiment scaling laws}, the hypothesis that increasing the number of training embodiments improves generalization to unseen ones, using robot locomotion as a test bed. We procedurally generate $\sim$1,000 embodiments with topological, geometric, and joint-level kinematic variations, and train policies on random subsets. We observe positive scaling trends supporting the hypothesis, and find that embodiment scaling enables substantially broader generalization than data scaling on fixed embodiments. Our best policy, trained on the full dataset, transfers zero-shot to novel embodiments in simulation and the real world, including the Unitree Go2 and H1. These results represent a step toward general embodied intelligence, with relevance to adaptive control for configurable robots, morphology co-design, and beyond.
\blfootnote{$^*$ Equal contribution. Correspondence addressed to Bo Ai $\langle$\texttt{bai@ucsd.edu}$\rangle$.} \vspace{-6pt}


\end{abstract}



\section{Introduction}

Two millennia ago, Heraclitus wrote that no one steps into the same river twice. Today, one might say that no agent acts with exactly the same body twice. Robotic embodiments change with injury, aging, tool use, manufacturing variation, and upgrades, and this variability will only grow as robots become more capable and widely deployed in the real world. Leveraging heterogeneous deployment data and understanding how to apply learned policies to novel, unseen embodiments would be instrumental to building generalist robots via data flywheels. However, it is unclear how to enable policies to transfer across a large number of distinct embodiments.  


Scaling has been a key driver of progress in deep learning, which can occur along multiple \textbf{dimensions}. Scaling \textbf{dataset size} and \textbf{model size} has improved generalization in vision~\citep{Kirillov2023Segment, Wen2024FoundationPose, Caron2021Emerging, Oquab2024Dinov2, Zhai2022Scaling, Sun2017revisiting, Tian2025diffusion, Mahajan2018exploring} and language~\citep{Ouyang2022Training, openai2024gpt4technicalreport, deepseekai2025deepseekr1, Kaplan2020Scaling, chowdhery2022palm, gao2021making, Hoffmann2022training, Ai2022whodunit, wu2023integrating, gao2025do}. In robotics, scaling the number of \textbf{tasks}~\citep{Fang2022Generalization, Kumar2023PreTraining, openx, Ghosh2024Octo, Brohan2023rt1, Zitkovich2023rt2, intelligence2025pi05, Fang2024RH20T, Kim2024openvla, black2410pi0} and \textbf{environments}~\citep{Ebert2022Bridge, Walke2023BridgeDataV2, Ghosh2024Octo, Lin2024DataScaling, Zitkovich2023rt2, Fang2020graspnet, intelligence2025pi05, gao2024intentionnet, bo2024invariance, Ai2022deep, shafiullah2023bringing, Mandlekar2023mimicgen, black2410pi0, Dasari2019Robonet} enable cross-task and cross-environment generalization. In this work, we explore a distinct and underexplored dimension of scaling: \textbf{robot embodiment}, the physical structure of robots. We hypothesize that scaling the number of training embodiments leads to better generalization to unseen embodiments, as the policies learn to capture shared control strategies across different physical structures. We conceptualize this hypothesized relationship as \textit{\textbf{embodiment scaling laws}}.

Studying this hypothesis requires addressing several open challenges. First, policy architectures need to (i) incorporate embodiment structure, (ii) adapt to varied observation and action spaces, and (iii) scale to large numbers of embodiments. Second, scaling experiments demand a large dataset of robot embodiments. We postulate that at least $\sim\!10^3$ embodiments are needed to reveal a glimpse of long-term trends, whereas existing literature is limited to $\sim\!10^1$ in the real world~\citep{openx, Ebert2022Bridge, Walke2023BridgeDataV2, Khazatsky2024droid, Kim2024openvla, black2410pi0, Dasari2019Robonet} and $\sim\!10^2$ in simulation~\citep{Patel2024GetZero}. Achieving this scale requires scalable and safe training and evaluation, which is only feasible in simulation at this point in time.

In simulation, we adopt a controlled setup to provide the first empirical validation of embodiment scaling laws. Proprioceptive locomotion serves as a foundational testbed: it has a small sim-to-real gap, depends primarily on morphology and dynamics, and avoids confounding factors from perception such as camera viewpoint, visual encoding, or rendering fidelity. We procedurally generate \ourdataset, a collection of $\sim$1,000 blueprint robot descriptions, spanning humanoids, quadrupeds, and hexapods, by varying topology, geometry, and joint-level kinematic constraints. To handle varied state and action spaces, we extend \gls{urma} \citep{bohlinger2024onepolicy} into a wider multi-head attention architecture. We adopt a two-stage learning framework \citep{Jia2022improving,Wan2023unidexgrasp} for scalable cross-embodiment learning: (i) train single-embodiment expert policies with \gls{rl}, and (ii) distill these experts into a single embodiment-aware \gls{urma} policy via behavior cloning (BC). By varying the number of embodiments used for BC, we quantify the effect of embodiment scaling on generalization to unseen embodiments. 

Overall, we present a large-scale empirical study of embodiment scaling laws across $\sim$1,000 robot embodiments. We design a general reward formulation, training curriculum, and domain randomization that enable scalable training of \gls{rl} expert policies without embodiment-specific tuning, accumulating a total of 2 trillion simulation steps. 
We observe positive scaling trends supporting the hypothesis, and find that embodiment scaling enables substantially broader generalization than data scaling on fixed embodiments.
The best policy, trained on 2 billion expert demonstration steps across the full set of training embodiments, zero-shot transfers to real-world robots, including the Unitree Go2 with varied kinematic constraints and the H1 humanoid. 
The findings underscore the potential of embodiment scaling for general embodied intelligence, and open up opportunities for embodiment-adaptive control, morphology co-design, and beyond.


\section{Related Work}

\textbf{Cross-embodiment generalization.} One goal of cross-embodiment learning is to enable control policies to generalize across robot embodiments without retraining.
Prior efforts often focus on transferring policies between a small number of robots by aligning dynamics, learning shared embeddings~\citep{zhu2024cross, chen2019learning}, or extracting transferable skills~\citep{hu2019skill, liumeta}. However, these methods are only able to transfer to a single or a few target embodiments.
Related work about scalable network architectures, such as graph neural networks~\citep{wang2018nervenet, Huang2020One} or Transformers~\citep{trabucco2022anymorph, furutasystem}, scale to more complex embodiments by conditioning on embodiment-specific information, but these works mostly use unrealistic and simplified robots that are not suitable for real-world transfer.
More recent approaches can be trained on a larger number of realistic robot embodiments, but they often rely on existing low-level controllers~\citep{shah2023gnm, song2024germ}, embodiment-specific decoders~\citep{doshi2024scaling}, other action abstractions~\citep{shafiee2024manyquadrupeds, eftekhar2024one}, or assume a fixed observation and action space \citep{Feng2022GenLoco}, limiting their generalization capabilities to pre-defined morphological structures.
\gls{urma}~\citep{bohlinger2024onepolicy} solves this issue by introducing a unified joint-level control architecture for arbitrary robot morphologies, but is validated only on 16 robots without studying scaling effects. 
Our work demonstrates broader cross-embodiment generalization than prior works by training a single policy on $\sim$1,000 embodiments, achieving zero-shot transfer to unseen embodiments in both simulation and the real world. 


\textbf{Robot locomotion.} In recent years, \gls{drl} has been applied to single embodiment robot locomotion to great success.
The combination of scalable on-policy \gls{rl} algorithms, such as \gls{ppo} \citep{schulman2017}, with fast and highly parallelizable simulators has enabled the training of powerful locomotion policies for quadruped \citep{miki2022,margolis2023,choi2023,caluwaerts2023,bridgethegap,zhuang2023,cheng2023} and humanoid robots \citep{siekmann2021,kumar2022,radosavovic2023,liao2024,zhuang2024}.
Techniques such as student-teacher learning \citep{chane2024,kaufmann2023champion}, curriculum learning \citep{kumar2021, rudin2022, margolis2022}, and domain randomization \citep{peng2018,campanaro2022,kumar2021} have enabled zero-shot sim-to-real transfer of these policies. 
Less data-hungry methods for learning directly on real robots, utilizing model-based or off-policy \gls{rl} algorithms \citep{smith2022,smith2023,levy2024,bohlinger2025gait}, and non-learning methods, such as \gls{mpc} \citep{jenelten2023,kasaei2021}, have also been proposed for legged locomotion, but generally trade their efficiency for worse performance with less robust gaits on challenging terrain or under strong perturbations.

\textbf{Robot embodiment generation.} 
Prior research in robot embodiment generation has pursued several directions. One prominent direction focuses on optimizing robot designs for specific tasks, where procedural and learning-based techniques generate embodiments tailored for enhanced performance in tasks such as locomotion~\citep{zhao2020robogrammar, azakami2022adversarial, rajani2023co} or manipulation~\citep{hazard2020automated}. Closer to our objectives is the use of embodiment generation to develop generalizable robot policies. Existing works have explored methods based on simplified kinematic trees~\citep{gupta2022metamorph, furutasystem}, randomization within a fixed morphology~\citep{Feng2022GenLoco}, diverse sensor configurations~\citep{eftekhar2024one}, or varied hand structures~\citep{patel2024get}.
However, these approaches are generally limited to a single robot class or topological template. In contrast, we introduce a comprehensive procedural generation framework that spans multiple morphology classes, including quadrupeds, hexapods, and humanoids, while varying topology, geometry, and kinematics for each of them. This enables a large-scale systematic study of embodiment scaling in locomotion.


\begin{figure*}[t]
    \centering
    \includegraphics[width=\textwidth]{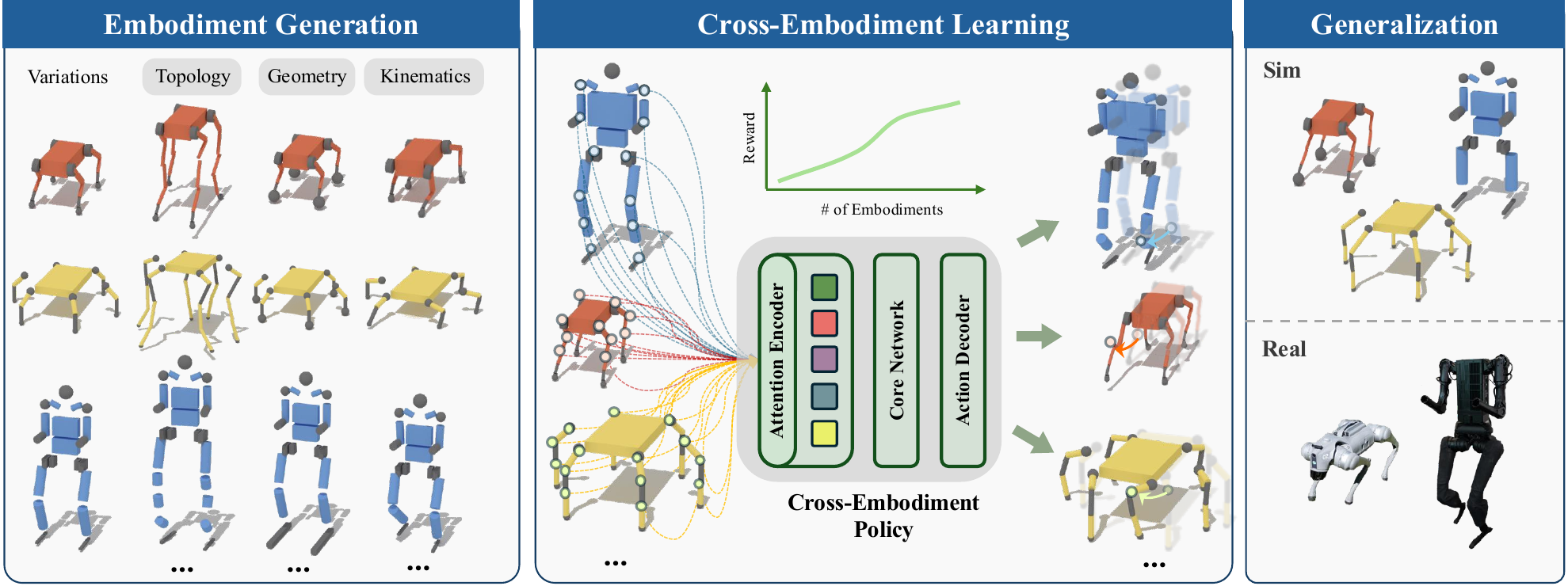} \vspace{-13pt}
    \caption{
        \textbf{Overview of our approach for studying embodiment scaling laws.} 
        We procedurally generate \ourdataset{}, a dataset of $\sim$1000 diverse robot embodiments with structured variations in topology, geometry, and kinematics. We train a single cross-embodiment policy using the \gls{urma} architecture, which handles varying observation and action spaces via attention-based joint encoding. We systematically vary the number of training embodiments to study how generalization scales with embodiment quantity. The policy trained on the full training dataset transfers zero-shot to novel simulated robots and real-world hardware with different morphologies.
    }
    \vspace{-10pt}
    \label{fig:framework}
\end{figure*}

\section{Methodology} \label{sec:method}


Generalizable cross-embodiment robot learning aims to train a control policy that can control diverse \textit{seen} and \textit{unseen} robot embodiments to solve a common task. Formally, let \(\mathcal{E}\) denote a set of embodiments sampled from $\mathcal{P}_{\mathcal{E}}$, where each embodiment \(e \in \mathcal{E}\) is defined as a triplet \( e = \langle \mathcal{G}, \mathcal{T}, \mathcal{K} \rangle \), where \( \mathcal{T} \) specifies the joint topology (\ie, number and connectivity), \( \mathcal{G} \) denotes link geometry (\eg, shape and size), and \( \mathcal{K} \) describes additional kinematic properties (\eg, joint types and range of motion). 
The control problem of each embodiment \(e\) is defined by a \gls{mdp} \(\mathcal{M}_e = \langle \mathcal{S}_e, \mathcal{A}_e, P_e, R_e, H \rangle\), where \(\mathcal{S}_e\), \(\mathcal{A}_e\), and \(P_e\) denote the state space, action space, and transition dynamics; \(R_e\) is the reward function; and \(H\) is the episode horizon. At any particular time step $t$, a policy predicts an action \(a_t \in \mathcal{A}_e\), conditioned on the robot state \(s_t \in \mathcal{S}_e\) and the embodiment descriptor \(\phi(e)\). In the specific case of robot locomotion, the policy is additionally conditioned on a x-y-yaw velocity command \(v_t \in \mathbb{R}^3\) with respect to the trunk frame, \ie, \(a_t \sim \pi(s_t, \phi(e), v_t)\).



During training, we optimize the policy to maximize the expected cumulative reward across training embodiments \(\mathcal{E}_{\text{train}} \subset \mathcal{E}\) with trajectories \(\tau = \{(s_0, a_0), \ldots, (s_H, a_H)\}\) sampled from \(\mathcal{M}_e\):
\setlength{\abovedisplayskip}{1pt}
\setlength{\belowdisplayskip}{1pt}
\begin{equation}
\pi^*_{\text{train}} = \arg\max_{\pi} \mathbb{E}_{e \in \mathcal{E}_{\text{train}}} \mathbb{E}_{\tau \sim \pi} \left[ \sum_{t=0}^{H} R_e(s_t, v_t, a_t) \right]. \label{eqn:train}
\end{equation}
The generalization performance is evaluated on a held-out set of embodiments
\(\mathcal{E}_{\text{test}} = \mathcal{E} \setminus \mathcal{E}_{\text{train}}\):
\setlength{\abovedisplayskip}{1pt}
\setlength{\belowdisplayskip}{1pt}
\begin{equation}
J_{\text{test}}(\pi^*_{\text{train}}) = \mathbb{E}_{e \in \mathcal{E}_{\text{test}}} \mathbb{E}_{\tau \sim \pi^*_{\text{train}}} \left[ \sum_{t=0}^{H} R_e(s_t, v_t, a_t) \right]. \label{eqn:test}
\end{equation}
We note that both \textit{learning} and \textit{generalizing} across embodiments present significant challenges. Differing observation and action spaces require policies to handle variable-sized and structurally different
inputs and outputs. Variations in kinematic constraints, self-collision profiles, and contact dynamics introduce embodiment-specific behaviors that complicate the optimization landscape of policy learning. Even further, generalizing to unseen embodiments demands that the policy captures meaningful shared control features that can be applied to novel physical embodiments. 

\textbf{Scaling hypothesis.} We hypothesize that generalization improves with the number of training embodiments, \ie, larger \(|\mathcal{E}_{\text{train}}|\) leads to higher \(J_{\text{test}}\). Intuitively, training on more diverse embodiments encourages the policy to extract structural features that transfer to novel robots. For instance, despite differences in leg length or joint placement, many embodiments share similar locomotion dynamics and constraints. Discovering a scaling trend would provide empirical support for an embodiment scaling law and offer actionable insights for building general-purpose control policies.

\textbf{Empirical setup.} To study the hypothesis, we fix a constant test set by randomly holding out 20\% of the generated embodiments. The remaining 80\% serve as the pool for constructing training subsets \(\mathcal{E}_{\text{train}}^{(i)} \subset \mathcal{E}_{\text{train}}\) at varying proportions \(i \in (0, 1]\). For each subset, we train a separate policy \(\pi^{(i)*}_{\text{train}}\) and evaluate it on the fixed \(\mathcal{E}_{\text{test}}\). This setup enables a systematic analysis of generalization performance \(J_{\text{test}}(\pi^{(i)*}_{\text{train}})\) as a function of training set size, probing for evidence of an embodiment scaling law.

Next, we describe how we generate diverse embodiments (\secref{sec:generation}), construct a policy to handle varying observation and action spaces (\secref{sec:model}), and train it on many embodiments (\secref{sec:training}).

\subsection{Embodiment Generation} \label{sec:generation}

We adopt a procedural generation pipeline to produce diverse robot embodiments spanning three commonly used morphology classes: humanoid~\citep{Cheng2024Expressive, kumar2022, Bjorck2025GROOT, Ji2024ExBody2, Sferrazza2024HumanoidBench, bohlinger2024onepolicy, shi2025toddlerbot}, quadruped~\citep{shafiee2024manyquadrupeds, kumar2021, Liu2024Visual, Yang2022Learning, He2024Agile, bohlinger2024onepolicy, Margolis2024RapidLocomotion, margolis2022, Tan2018simtoreal, cheng2023}, and hexapod~\citep{bohlinger2024onepolicy, zhang_development_2014, zang_perceptive_2023, Qu2024versatile, Ouyang2021AdaptiveLocomotion, azayev_blind_2020, chiu_development_2020, haitao_yu_cpg-based_2013}.
Our generated robots follow common design patterns using realistic base components, such as link shapes, dimensions, and motor properties, but are procedurally composed into novel embodiments by varying their parameters. Geometric variation is introduced by scaling individual links and overall body size. Topological variation is achieved by changing the number of knee joints per leg within each morphology class. We also vary joint limits to implement kinematic variations.  
In total, we generate 1,012 distinct robots, including 348 humanoids, 332 quadrupeds, and 332 hexapods, to form the \ourdataset dataset (\figref{fig:teaser}).
Our resulting dataset is diverse in various aspects, as reflected in post-generation statistics (\figref{fig:dataset-statistics}). 
More details about the generation process are provided in \appendref{appendix:generation}.

\begin{figure*}[t]
    \centering
    \includegraphics[width=\linewidth]{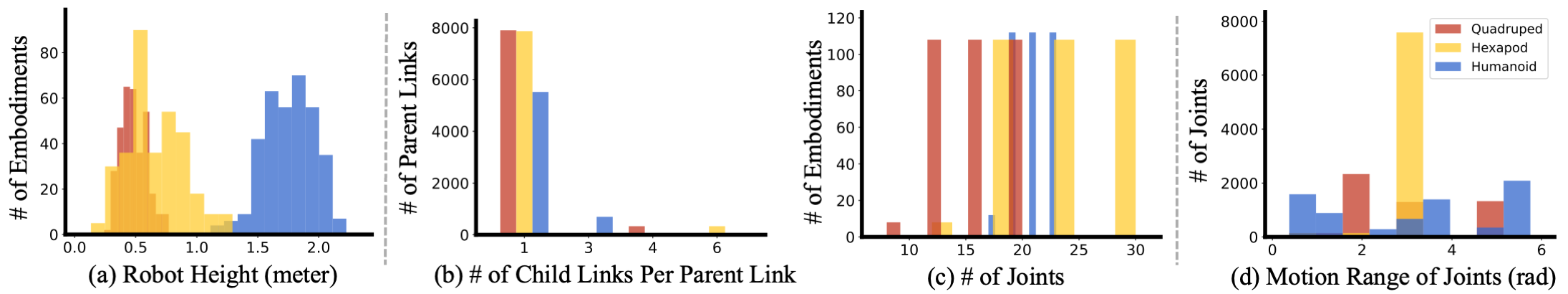}
    \caption{\textbf{Empirical distributions of embodiment variations in \ourdataset{}.}
    The statistics reflect geometric~(a), topological~(b,c), and kinematic~(d) variability of embodiments in our dataset. 
    }
    \vspace{-10pt}
    \label{fig:dataset-statistics}
\end{figure*}

\subsection{Cross-Embodiment Policy Architecture} \label{sec:model}
To train a policy that can control $\sim$1000 different embodiments with different state and action spaces, we use \gls{urma}, an embodiment-aware architecture for robots with arbitrary numbers of joints~\citep{bohlinger2024onepolicy}.
\gls{urma} handles the differently sized partially observable states (observations) $o$ of different embodiments by splitting them into fixed-length general observations $o_g$ and varying-length joint-specific observations $o_j$, depending on the set of joints $J$ of the current embodiment $e$ (i.e., $o = (o_g,\{o_j\}_{j\in J})$).
The embodiment descriptors $\phi(e)$ are used to generate joint description vectors $d_j$, which can uniquely describe every joint of the embodiment and are made up of the fixed dynamics and kinematics properties of the joint and its underlying motor.
The joint-specific observations are processed by an attention encoder and are summed up into the joint latent vector
\setlength{\abovedisplayskip}{0pt}
\setlength{\belowdisplayskip}{0pt}
\begin{equation}
    \bar{z}_\text{joints} = \sum_{j \in J} z_j, \quad \quad z_j = \frac{\exp\left(f_\phi(d_j) / \tau \right)}{\sum_{L_d} \exp\left(f_\phi(d_j) / \tau\right)} f_\psi(o_j),
\end{equation}
where $f_\phi$ (with latent dimension $L_d$) and $f_\psi$ are the encoders for the joint descriptions and joint observations, respectively, and $\tau$ is the learnable temperature parameter of the softmax. Intuitively, the attention mechanism fuses joint observations based on their descriptions so that $\bar{z}_\text{joints}$ has global information about the embodiment. 
The encoded joint latent vector is then concatenated with the general observations and processed by a core network to generate an action latent vector $\bar{z}_\text{action} = h_\theta(o_g, \bar{z}_\text{joints})$.
To handle the differently sized action spaces, \gls{urma} concatenates the action latent vector with each encoded joint description vector in batch to decode a single action for each joint:
\begin{equation}
    a_j = \mu_\nu(g_\omega(d_j), \bar{z}_\text{action}, z_j),
\end{equation}
where $g_\omega$ is the action encoder for the joint descriptions, $\mu_\nu$ is the final action decoder. In our work, we incorporate multi-head attention \citep{Vaswani2017attention} into \gls{urma}, enabling the policy to attend to different joint-level features in parallel and better capture complex inter-joint dependencies (\appendref{appendix:urma-architecture}).

\subsection{Two-Stage Policy Learning}  \label{sec:training}

To scale cross-embodiment policy learning to a large number of robots, we adopt a two-stage paradigm. First, we train embodiment-specific expert policies using standard \gls{rl}. Then, we collect demonstration data from these experts and train a single student policy via imitation learning, conditioned on embodiment descriptors. This approach allows learning across $\sim$1000 robots while maintaining tractable memory usage and stable training dynamics. The setup also mirrors real-world data pipelines, where large datasets are collected offline and reused across training runs (\eg, \citep{openx}).

\textbf{Expert training.}
We develop a unified \gls{rl} locomotion training pipeline applicable to all embodiments with minimal tuning. Key components include extensive domain randomization, performance-based curriculum learning, and regularization terms that encourage stable and natural locomotion (e.g., penalizing jittering movements and excessive ground contact). All robots in one morphology class share one set of hyperparameters for scalable training. 

Training robust policies for $\sim$1,000 robot embodiments is computationally demanding. We use NVIDIA Isaac Lab~\citep{mittal2023} to train single-embodiment policies across 4,096 parallel environments with \gls{ppo}~\citep{schulman2017}.
Training all experts takes approximately 5 days on 160 NVIDIA RTX 4090/3090 GPUs, totaling over 2 trillion simulation steps. Full training details are 
provided in \appendref{appendix:rl_training}.

\textbf{Student distillation.}
Given expert policies \(\{\pi_e\}_{e \in \mathcal{E}_{train}}\), we collect a demonstration dataset by rolling out each policy for 600 timesteps in 4,096 parallel environments, totaling 2 billion samples across all embodiments. We then train \gls{urma} by minimizing the \gls{mse}: 
\begin{equation}
\mathcal{L}_{\text{BC}} = \mathbb{E}_{(s_t, e, a_t) \sim \mathcal{D}} \left[ \left\| \pi(s_t, \phi(e)) - a_t \right\|^2 \right],
\end{equation}
where \(\mathcal{D}\) is the expert demonstration dataset. The student policy conditions on the embodiment descriptor \(\phi(e)\), enabling it to generalize across the generated embodiments with different geometry, topology, and kinematics. Training the model on the full demonstration dataset takes one week using a NVIDIA H100 GPU. More details about the distillation process can be found in \appendref{appendix:distillation}.


\section{Experiments} \label{sec:exp}

In this section, we 
conduct a large-scale empirical study to investigate the scaling behavior of cross-embodiment learning. 
Our experiments are designed to answer the following key research questions:
\begin{itemize}[nosep]
    \item[\textbf{Q1.}] How does the generalization performance of the cross-embodiment policy scale with the number of training embodiments? (Sec.~\ref{sec:scaling_results})
    \item[\textbf{Q2.}] Can the learned policy generalize zero-shot to unseen embodiments, including real-world robots, and handle varied kinematic constraints? (Sec.~\ref{sec:real_world_results})
    \item[\textbf{Q3.}] Does the policy network learn meaningful, structured representations of the space of robot embodiments and morphologies through cross-embodiment training? (Sec.~\ref{sec:representation_results})
\end{itemize}


\subsection{Studying Embodiment Scaling Laws}
\label{sec:scaling_results}

\begin{figure*}[t]
    \centering
    \setlength{\tabcolsep}{1pt} 
    \renewcommand{\arraystretch}{0.5} 
    \includegraphics[width=\linewidth]{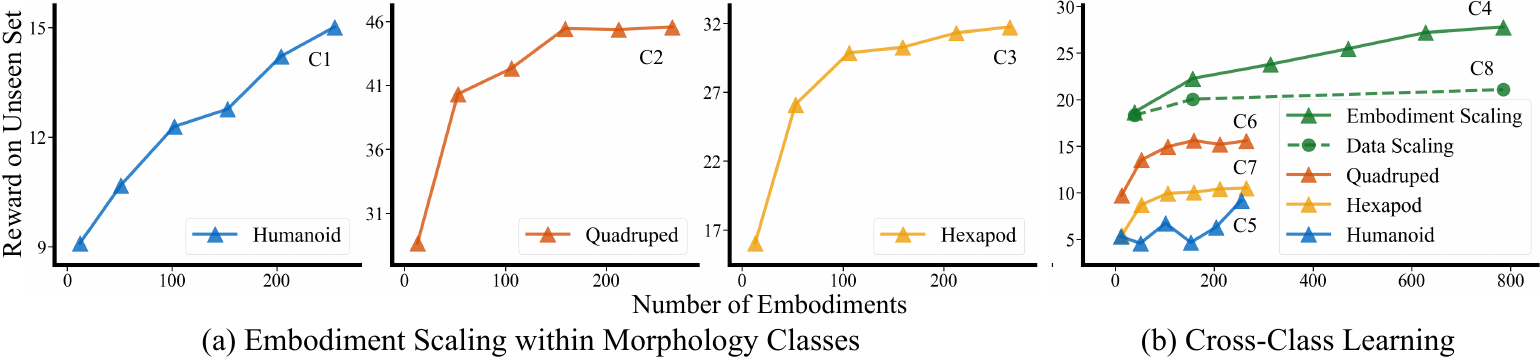} \vspace{-15pt}
    \caption{\textbf{Results on embodiment scaling.} We evaluate generalization performance as a function of the number of training embodiments. (a) In-class study: policies are trained and tested within the same morphology class (humanoid, quadruped, or hexapod). (b) Cross-class study: We train policies on the full training set (green) and compare performance against policies trained on only the individual classes, while all policies are evaluated on the test set containing all classes. The proportion of training embodiments ($i \in \{0.05, 0.2, 0.4, 0.6, 0.8, 1.0\}$) is denoted in the x-axis. While the underlying reward function is the same, the reward scales differ across classes due to inherent differences in the embodiments (e.g., humanoids are less stable than quadrupeds) and unnormalized reward formulations (e.g., humanoids experience larger ground contact forces).}
    \label{fig:curves}
    \vspace{-10pt}
\end{figure*}


We train and evaluate our policies under multiple setups, with results illustrated in \figref{fig:curves}. We analyze generalization patterns from three perspectives below. Additional results on out-of-distribution generalization are provided in Appendix~\ref{appendix:additional_ood}.

\textbf{Scaling within each embodiment class.}
We conduct training and evaluation separately for each morphology class (humanoid, quadruped, and hexapod) in \ourdataset, resulting in curves \curve{1}–\curve{3}. For each morphology class, we observe a clear scaling trend: increasing the number of training embodiments from 0.05 to 1.0 can double the cumulative reward. The rate of convergence varies by class: for quadrupeds and hexapods, performance saturates around 100 training embodiments, while for humanoids, it continues to improve steadily with more training data, likely due to greater instability and control difficulty. This suggests that more challenging embodiments may benefit more from larger-scale embodiment scaling.

\textbf{Scaling across embodiment classes.}
We train on the full combined dataset of all three classes and evaluate on a unified test set (\curve{4}). The resulting curve begins at a reward of 18 and rises consistently to nearly 30, demonstrating that scaling across diverse embodiments enables broader generalization. We further evaluate the policies trained on individual morphology classes (corresponding to \curve{1}–\curve{3}) on the combined test set, obtaining (\curve{5}–\curve{7}). Since each of these models has only seen a single morphology class during training, their performance on the mixed test set is limited. In contrast, the best point on \curve{4} achieves 2–5$\times$ higher average reward than \curve{5}–\curve{7}, demonstrating that training across diverse morphology classes enables substantially broader generalization.


\textbf{Comparison with pure data scaling.}
To disentangle the effects of embodiment diversity from data quantity, we collect a dataset using only 5\% of the training embodiments and vary the number of trajectories per embodiment for distillation (\curve{8}). We find that performance quickly saturates: the policy nearly reaches its peak at 0.2 data scale (4$\times$ data as 0.05), with negligible gains beyond that. This highlights that, if the goal is to achieve broad embodiment-level generalization, it is ineffective to only increase data volume on a small set of embodiments. Embodiment scaling is essential.


\begin{figure*}[t]
    \centering
    \includegraphics[width=\linewidth]{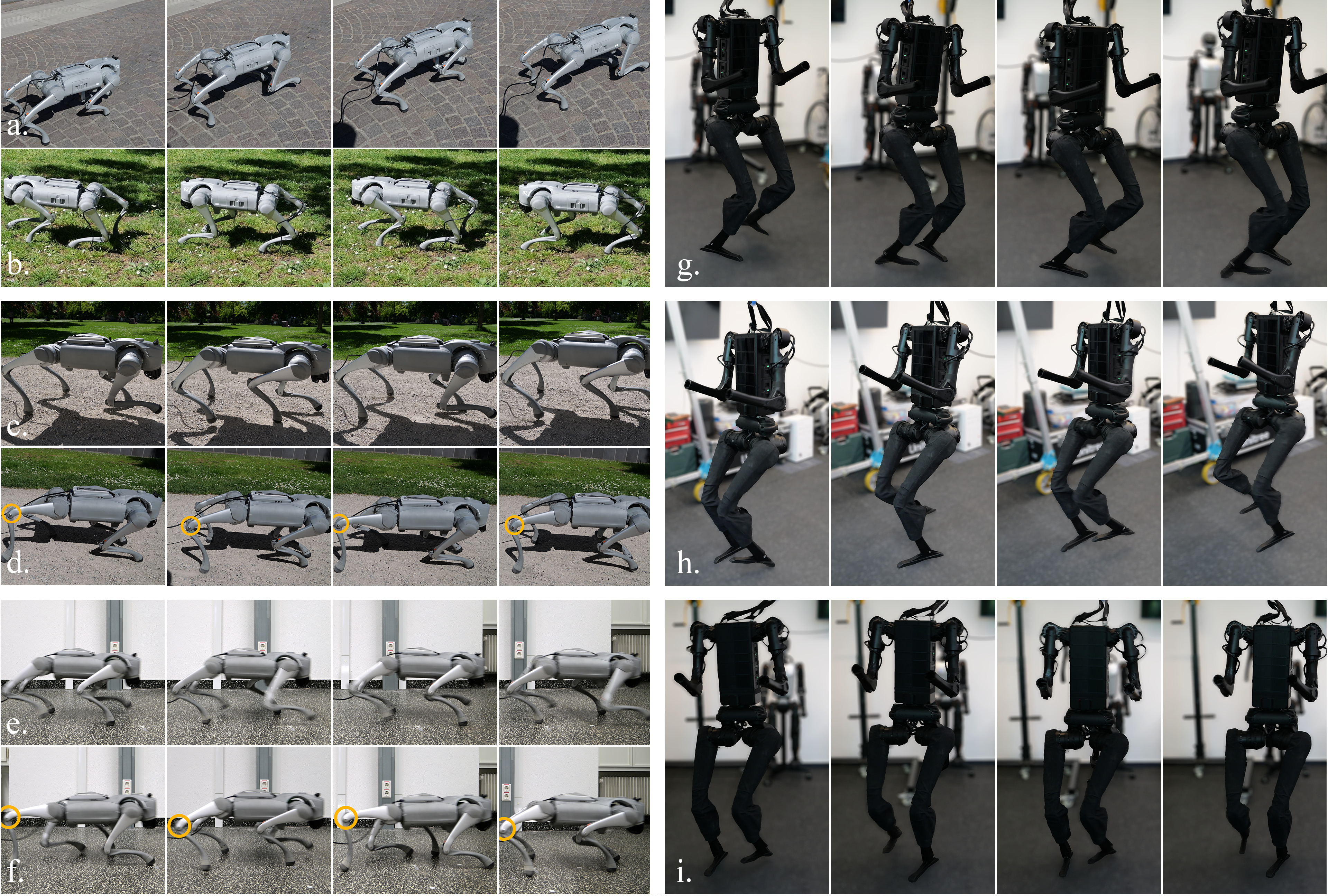}
    \caption{
        \textbf{Zero-shot generalization to unseen real-world robots.}
        Our \gls{urma} policy, trained on 817 diverse simulated embodiments, successfully transfers zero-shot to control the Unitree Go2 quadruped and Unitree H1 humanoid in the real world.
        \textbf{a, b:} The policy can perform forward and backward locomotion on cobblestone and grass terrain with the Go2.
        \textbf{c, d, e, f:} We test the policy's adaptation to kinematic constraints by artificially restricted the joint limits on the right rear knee of the Go2 by 20\%.
        The policy effectively compensates for the limited range of motion, resulting in a stable limping gait on gravel (d) and indoors (f), compared to the unrestricted gait (c, e).
        \textbf{g, h:} Zero-shot transfer on the H1 works well in a lab environment, showing decent forward and backward locomotion.
        \textbf{i:} Walking side-to-side with H1 is slower as in simulation but stable in the real world. 
   }
    \vspace{-10pt}
    \label{fig:real-world}
\end{figure*}

\subsection{Real-World Generalization Test}
\label{sec:real_world_results}

To validate real-world transfer capabilities, we conducted zero-shot deployments of our best-performing policy, trained on the full training set of 817 simulated embodiments, on two real robots: the Unitree Go2 quadruped and the Unitree H1 humanoid, neither of which was included in the training set $\mathcal{E}_{\text{train}}$, although robots with similar kinematic structures were present.

Figure~\ref{fig:real-world} shows the policy successfully generalizing to the two real robots without any fine-tuning or modifications, using only the URDF of the respective robot to generate the embodiment descriptor $\phi(e)$.
The Go2 demonstrated robust and stable walking gaits across diverse terrains such as grass, cobblestone, and gravel (a-c). Similarly, the H1 was able to maintain stable locomotion, tracking the desired velocity commands while walking on flat ground with rubber mats in a lab environment (g-i). While the transfer worked for both robots, the policy transferred worse to the H1 compared to the Go2, highlighting the need for potentially even more diverse humanoid robots in the training set.

To probe the policy's ability to handle kinematic variations in the real world, we artificially restricted the joint limits of the knee joints of the Go2 in 12 different configurations. We enforce different joint limits by pushing towards the artificial limits with high gains when the joint angles exceed them.
Figure~\ref{fig:real-world} (d, f) shows that the policy was able to transfer the adaptations it learned in simulation as it keeps the restricted rear right leg further back and maintains a stable limping gait.


\vspace{-5pt}
\subsection{Understanding Learned Embodiment Representations}
\label{sec:representation_results}
\vspace{-5pt}

\begin{figure}[t]
    \centering
    \includegraphics[width=\columnwidth]{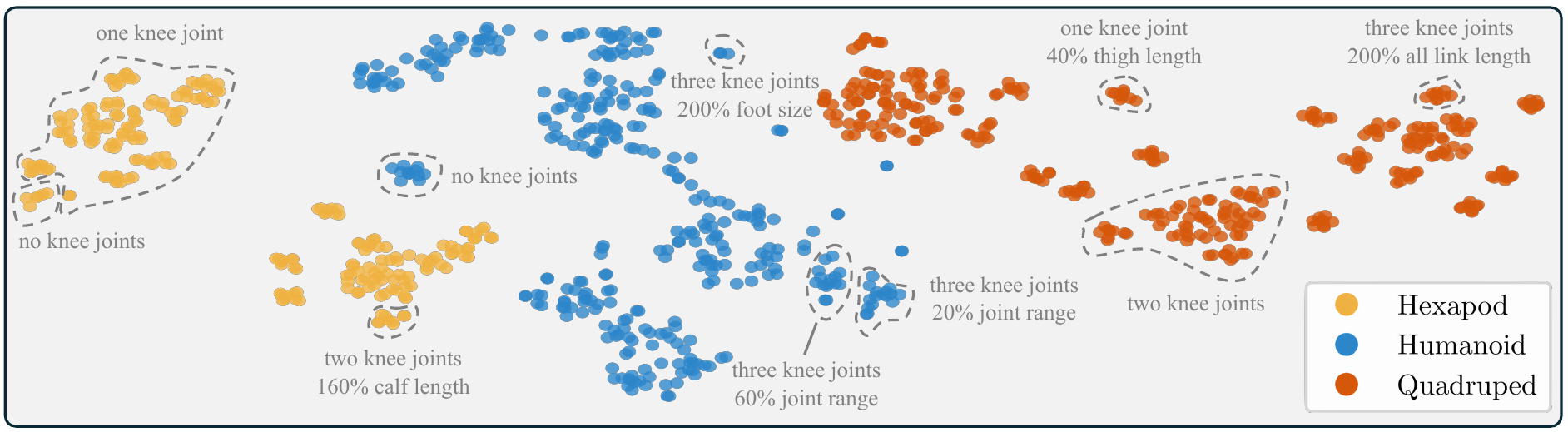}
    \caption{
        \textbf{Visualization of the embodiment latent space.}
        \gls{tsne} projection of the action latent vectors on the complete \ourdataset dataset from the \gls{urma} policy trained on the full training set.
        Points are colored by morphology class. The clear clustering based on morphology class, and finer sub-structures related to the number of knee joints or kinematic and geometric properties, suggest that the policy learns a meaningful and structured representation of diverse embodiments, capturing functional similarities that help during cross-embodiment learning and enable generalization.
    }
    \vspace{-10pt}
    \label{fig:clustering}
\end{figure}

To gain insight into the internal representations learned by our policy, we performed a \gls{tsne}~\citep{vandermaaten08a} analysis on the action latent vectors $\bar{z}_\text{action}$ produced by \gls{urma} for each embodiment.
Figure~\ref{fig:clustering} shows that the learned representations naturally cluster according to the robot morphology, clearly distinguishing humanoids, quadrupeds, and hexapods.
For all three morphologies, large clusters around the number of knee joints separate most of the latent space, showing the impact of additional joints on the policy.
Many finer sub-clusters emerge based on different geometric and kinematic variations for a given number of knee joints.
This structured representation indicates that our policy captures meaningful embodiment-specific features that generalize, mostly, within the morphology classes, whereas patterns across classes are less clear.
Additional visualizations using PCA \citep{Jolliffe2012Principal} and UMAP \citep{McInnes2018UMAP} can be found in \appendref{appendix:further_latent_analysis}.

\vspace{-5pt}
\section{Conclusion} \label{sec:conclusion}
\vspace{-5pt}


We conceptualize embodiment scaling laws and provide preliminary empirical evidence through a large-scale study on robot locomotion, using a procedurally generated dataset \ourdataset{}.
Our results show that increasing the number of training embodiments improves generalization to unseen ones, with more challenging morphologies benefiting from continued scaling. Scaling across embodiment classes further enhances generalization, while simply increasing data volume on a fixed set of robots yields diminishing returns. We also demonstrate successful sim-to-real transfer of the learned cross-embodiment policy. As robotic platforms grow more diverse, the ability to learn from and generalize across embodiments becomes increasingly critical. We hope this work offers a step toward general-purpose cross-embodiment intelligence. 


\clearpage

\section*{Limitations} \label{sec:limitations}

While our work provides empirical evidence for embodiment scaling laws in robot learning, several limitations remain.  

First, regarding task setup, our study focuses exclusively on locomotion on flat terrain for a highly controlled, foundational study. Extending this analysis to more complex tasks, such as vision-based manipulation or loco-manipulation, is an interesting exploration for future work. 

Second, although our procedural generation pipeline produces a wide range of embodiments varying in topology, geometry, and joint constraints, it does not exhaustively cover the design space. Several factors, including body mass distribution, joint damping, and actuation type, are held fixed within a morphology class. Expanding the generation space to include such parameters could yield more robust generalization and offer further insights into the scaling behavior.  


Finally, real-world experiments are limited to two physical robot platforms. Although we have modified their joint limits to create more kinematic variations, and existing results already demonstrated promising zero-shot transferability, broader validation on more diverse physical platforms, such as modular or reconfigurable robots, would provide stronger support for the generality of our findings. 

Despite these limitations, we believe our study represents an important step toward understanding embodiment-level generalization and highlights the key role of embodiment scaling in the pursuit of generalizable robot learning.


\acknowledgments{
This research is funded by the NSF AI-Center TILOS, the Hillbot Embodied AI Fund, the National Science Centre Poland (Weave programme UMO-2021/43/I/ST6/02711), and by the German Science Foundation (DFG) (grant number PE 2315/17-1).

Co-author Hao Su is the CTO for Hillbot and receives income. The terms of this arrangement have been reviewed and approved by the University of California, San Diego, in accordance with its conflict of interest policies.

We thank the German Research Center for AI (DFKI), Research Department: Systems AI for Robot Learning, for lending the Unitree Go2 and Unitree H1 robots.

Finally, we thank Oleg Kaidanov (DFKI, TU Darmstadt) for his continuous help with the real-world robot deployment, and we are grateful to the UCSD Su Lab members for facilitating extended access to compute resources, which made large-scale experiments feasible.
}


\bibliography{references}  

\clearpage
\appendix
\section*{Appendix}

\section{Expert Training} \label{appendix:rl_training}

\subsection{Observation and Action Space}
\label{sec:obs_action_space}
The observation space of the expert policies includes the joint angles, joint velocities, previous actions, trunk angular velocities, gravity vector and the command velocities.
The observation space of the critics includes the same observations as for the policies, but also includes privileged information: the trunk linear velocity, trunk height over the ground, feet contact states and feet air times.

The policies control the robots at 50 Hz with a PD controller, where the target joint angles are generated by scaling the action of the policy and adding it to the nominal joint configuration of the robot: $q_{\text{target}} = q_{\text{nominal}} + \sigma \cdot a$.
We define the nominal joint configuration as a standing pose of a robot and use the same configuration for all robots of the same morphology class (see \appendref{sec:nominal_joint_config}).
For the action scaling factor $\sigma$, we use $0.3$ for quadrupeds and hexapods, and $0.75$ for humanoids.
For the PD controller, we use $K_p = 20$ and $K_d = 0.5$ for quadrupeds, $K_p = 25$ and $K_d = 0.5$ for hexapods, and $K_p = 60$ and $K_d = 2.0$ for humanoids.

\subsection{Domain Randomization}
\label{sec:domain_randomization}
To enable sim-to-real transfer of the trained policies, we add strong domain randomization during training. We use a performance-based curriculum learning approach, where the domain randomization ranges are increased from 0 (or their mean if not zero-centered) to the final values in Table \ref{tab:domain_randomization} over the course of training.
This curriculum approach allows the policy to learn basic locomotion first in the simplest possible environment before adapting to wider variations.
We define a curriculum coefficient from 0 to 1, which is multiplied with the domain randomization ranges (and the reward penalty coefficients).
The coefficient of an environment is increased by $0.01$ if the policy completed the episode without falling, and the average tracking error of the target x,y velocity is below $0.4$~m/s, and the coefficient is reduced by $0.01$ otherwise.

Every embodiment in \ourdataset uses the same domain randomization ranges.
The "starting" values (naming scheme in Table \ref{tab:domain_randomization})  are sampled uniformly at the start of every episode to randomize the starting state of the robot.
The "noise" values are sampled uniformly for every simulation step to add noise to the observations.
The values of every other parameter are sampled uniformly every simulation step with a probability of $0.002$ (on average every 500 steps / every 10 seconds).
Pushes are applied as linear velocities to the trunk of the robot.

\begin{table}[t]
\centering
\def\arraystretch{1.3}
\caption{
\textbf{Domain randomization configuration.}
Domain randomization values and ranges for every randomized parameter during the expert \gls{rl} training. The values in the table are the maximum values and ranges in the curriculum when reaching the final curriculum coefficient of 1.  
}
\begin{tabular}{l l}
\toprule
Parameter & Value \\
\hline
Max action delay & 1 \\
Chance for action delay & 0.05 \\
Min \& max motor strength & (0.5, 1.5) \\
Min \& max P gain factor & (0.5, 1.5) \\
Min \& max D gain factor & (0.5, 1.5) \\
Min \& max joint position offset & (-0.05, 0.05) \\
Min \& max starting orientation factor & (-0.0625, 0.0625) \\
Min \& max starting joint position factor & (-0.5, 0.5) \\
Min \& max starting joint velocity factor & (-0.5, 0.5) \\
Min \& max starting linear velocity & (-0.5, 0.5) \\
Min \& max starting angular velocity & (-0.5, 0.5) \\
Joint position noise & 0.01 \\
Joint velocity noise & 1.5 \\
Angular velocity noise & 0.2 \\
Gravity velocity noise & 0.05 \\
Joint observation dropout chance & 0.05 \\
Min \& max static friction & (0.05, 2.0) \\
Min \& max dynamic friction & (0.05, 1.5) \\
Min \& max restitution & (0.0, 1.0) \\
Min \& max added mass & (-2.0, 2.0) \\
Min \& max gravity & (-8.81, 10.81) \\
Min \& max joint friction & (0.0, 0.01) \\
Min \& max joint armature & (0.0, 0.01) \\
Min \& max pushes in x & (-1.0, 1.0) \\
Min \& max pushes in y & (-1.0, 1.0) \\
Min \& max pushes in z & (-1.0, 1.0) \\
\bottomrule
\end{tabular}
\label{tab:domain_randomization}
\end{table}

\clearpage

\subsection{Reward Function}

Table \ref{tab:reward_terms} contains all reward terms and coefficients of the reward function for the expert training of all robots in the \ourdataset dataset.
Joint-based (T6-T12) and feet-based (T14-T17) reward terms are calculated as the mean over every joint and foot, respectively, to account for the varying amounts of joints and feet of the generated embodiments.
The coefficients of all penalties are attached to the curriculum coefficient (see \appendref{sec:domain_randomization}) and thus linearly increase from 0 to the final values in Table \ref{tab:reward_terms} over the course of training.
This makes the training process less sensitive to the precise values of the coefficients.

\begin{table}[h]
\centering
\def\arraystretch{1.3}
\caption{
\textbf{Reward terms for the \gls{rl} training of embodiment-specific experts.}
All reward terms and the corresponding coefficients that compose the reward function for the expert training.
While all the coefficients work for all embodiments, for the final experiments, we tweaked four coefficients for the humanoid embodiments to improve the style of the gait: \textsuperscript{*1} $3.0$, \textsuperscript{*2} $1.5$, \textsuperscript{*3} $43.2$, \textsuperscript{*4} $6$e-$3$. 
}
\begin{tabular}{l l l l}
\toprule
& Term & Equation & Coefficient \\
\hline
T1 & Xy velocity tracking & ${\exp(- |v_{xy} - c_{xy}|^2 / 0.25)}$ & $2.0$ \textsuperscript{*1} \\
T2 & Yaw velocity tracking & ${\exp(- |\omega_{\text{yaw}} - c_{\text{yaw}}|^2 / 0.25)}$ & $1.0$ \textsuperscript{*2} \\
T3 & Z velocity penalty & ${-|v_z|^2}$ & $2.0$ \\
T4 & Pitch-roll velocity penalty & ${-|\omega_{\text{pitch, roll}}|^2}$ & $0.05$ \\
T5 & Pitch-roll position penalty & ${-|\theta_{\text{pitch, roll}}|^2}$ & $5.0$ \\
T6 & Joint nominal differences penalty & ${-|q - q^{\text{nominal}}|^2}$ & $14.4$ \textsuperscript{*3} \\
T7 & Joint position limits penalty & ${-\bar{\bbone}(0.9 q_{\text{min}} < q < 0.9 q_{\text{max}})}$ & $120.0$ \\
T8 & Joint velocity limits penalty & ${-\bar{\bbone}(0.9 \dot{q}_{\text{min}} < \dot{q} < 0.9 \dot{q}_{\text{max}})}$ & $10.0$ \\
T9 & Joint accelerations penalty & ${-|\ddot{q}|^2}$ & $5$e-$6$ \\
T10 & Joint torques penalty & ${-|\tau|^2}$ & $2.4$e-$4$ \\
T11 & Action rate penalty & ${-|a_t - a_{t-1}|^2}$ & $0.12$ \\
T12 & Action smoothness penalty & ${-|a_t - 2 a_{t-1} + a_{t-2}|^2}$ & $0.12$ \\
T13 & Walking height penalty & ${-|h - h_{\text{nominal}}|^2}$ & $30.0$ \\
T14 & Air time penalty & ${-\textstyle{\sum_f} \bbone(p_f) (p_f^T - 0.5)}$ & $0.1$ \\
T15 & Symmetry penalty & ${-\textstyle{\sum_f} \bar{\bbone}(p_f^{\text{left}}) \bar{\bbone}(p_f^{\text{right}})}$ & $0.5$ \\
T16 & Feet y distance penalty & ${-|f^{\text{actual}}_\text{y distance} - f^{\text{target}}_\text{y distance}|^2}$ & $2.0$ \\
T17 & Feet force penalty & ${-|f_\text{force}|^2}$ & $8$e-$3$ \textsuperscript{*4} \\
T18 & Self-collision penalty & ${-\bbone_\text{self-collision}}$ & $1.0$ \\
\bottomrule
\end{tabular}
\label{tab:reward_terms}
\end{table}
\clearpage


\subsection{PPO Hyperparameters}
We use the same \gls{ppo} hyperparameters for the training of all expert policies, detailed in Table \ref{tab:hyperparams}.
Searching for better hyperparameters for every embodiment might lead to increased performance but is impractical when considering training $\sim$1000 embodiments.
The chosen hyperparameters are based on common practices in legged locomotion research~\citep{rudin2022,bohlinger2024onepolicy} and preliminary tuning on a small subset of embodiments.

\begin{table}[h]
\centering
\def\arraystretch{1.3}
\caption{
\textbf{\gls{ppo} hyperparameters for expert policy training. }
}
\begin{tabular}{l l}
\toprule
Hyperparameter & Value \\
\hline
Batch size & 98304 \\
Mini-batch size & 24576 \\
\# epochs & 5 \\
Initial learning rate & 0.001 \\
Learning rate schedule & Adaptive with target KL 0.01 \\
Entropy coefficient & 0.002 \\
Discount factor & 0.99 \\
GAE $\lambda$ & 0.95 \\
Clip range & 0.2 \\
Max gradient norm & 1.0 \\
Initial action standard deviation & 1.0 \\
Clip range action mean & -10.0, 10.0 \\
Policy and critic hidden layers & [512, 256, 128] \\
Activation function & ELU \\
\# training iterations & 17500 (quadruped, hexapod), 42500 (humanoid) \\
\bottomrule
\end{tabular}
\label{tab:hyperparams}
\end{table}

\clearpage


\section{Embodiment Generation} \label{appendix:generation}
\subsection{Basic Units for Quadruped, Humanoid and Hexapod}
\label{appendix:basic_units}


Tables~\ref{tab:basic_units} and~\ref{tab:joint_limits} provide the base values for geometry-related and kinematics-related parameters, respectively, for representative links across quadruped, humanoid, and hexapod morphologies. For the humanoid class, we report parameters for the trunk and the left-side lower-body links. For quadruped and hexapod classes, we include the front-left leg. The remaining components are either symmetric or peripheral to locomotion (e.g., arms or head for humanoids) and therefore omitted.

The base parameter values are partially inspired by the Unitree Go2 and H1 platforms, offering a degree of realism without exact replication. This design choice is consistent with prior work such as GenLoco~\citep{Feng2022GenLoco}, which abstracts physical characteristics from real robots to define a diverse yet grounded design space. Robots instantiated with these values correspond to a $1.0\times$ variation setting (\ie, no geometric, kinematic, or topological scaling applied), and serve as the reference point for applying the variation factors listed in Tables~\ref{tab:basic_units} and~\ref{tab:joint_limits}.

To support meaningful evaluation of generalization, these reference robots are excluded from the training set. Every robot in the training set differs from Go2 and H1 by at least one geometric, topological, or kinematic variation, along with additional discrepancies due to loose alignment in parameter values (\eg, each joint in the humanoid closest to H1 differs by a few centimeters, and the overall height differs by approximately 10 cm). This diversity encourages the learned policy to capture broadly transferable motion patterns. As discussed in \secref{sec:exp}, empirical results suggest that the policy has acquired sufficiently generalizable behaviors to support both cross-embodiment and sim-to-real transfer, which is generally considered highly challenging. 



\subsection{Generation Algorithm}
We construct each robot embodiment in a tree-like structure by iteratively connecting links using joints, following the URDF specification and the basic units described in Section~\ref{appendix:basic_units}. The construction procedure varies slightly across morphologies:
\begin{itemize}
    \item \textbf{Humanoids}: The root node is the pelvis. We first append the torso and hip links, then attach the shoulder and arm links for the upper body, followed by the thigh, calf, and foot links for the lower body.
    \item \textbf{Quadrupeds and hexapods}: The root node is the trunk. We sequentially append the hip links to the trunk, then connect the leg and foot links to form the complete body.
\end{itemize}



To ensure diversity in the generated embodiments, we introduce variations in geometry, topology, and kinematics during the construction process, as detailed in \secref{sec:method}. Table~\ref{tab:gen_variations} summarizes the variation parameters and their corresponding candidate values. While most parameters are self-explanatory, we clarify a few specific cases:

\begin{itemize}
    \item \textbf{Number of knee joints}: If a leg is configured with zero knee joints, the calf link is omitted, and the thigh link is directly connected to the foot.
    \item \textbf{Foot link size}: For humanoids, foot links are modeled as boxes and scaled by length; for quadrupeds and hexapods, foot links are modeled as spheres and scaled by radius.
    \item \textbf{Joint limit variation}: Joint limits are varied by uniformly scaling the nominal joint ranges about the nominal joint position, which serves as a fixed point.
\end{itemize}


\subsection{Nominal Joint Configurations}
\label{sec:nominal_joint_config}

Nominal joint configurations are used to initialize robot poses during training, contribute to reward terms that discourage deviations too far from these default joint angles, and function as offsets to the actions of the expert and distillation policies. As such, they serve as useful regularizers for learning realistic and efficient gaits. To support scalability across diverse morphologies, we generate nominal configurations by reusing unit values across the generated embodiments. The nominal joint angles used are summarized in \tabref{tab:nominal_joint_config}.



\begin{table}[h]
  \centering
  \renewcommand{\arraystretch}{1.3}
  \caption{\textbf{Base geometry and mass parameters for representative link types in the embodiment generation pipeline used in \ourdataset{}.} 
    Geometry dimensions are specified according to shape type: Sphere (radius), Cylinder (length, radius), and Box (length, width, height).
    }
  \begin{tabular}{l l l l r}
    \toprule
    Class & Link Name                & Geometry Type & Geometry Dimension (m)            & Mass (kg)  \\
    \midrule
    Humanoid   & Pelvis                   & Sphere        & (0.05,)                   & 5.390  \\
               & Torso                    & Box           & (0.08, 0.26, 0.18)        & 17.789 \\
               & Hip yaw link    & Cylinder      & (0.02, 0.01)              & 2.244  \\
               & Hip roll link    & Cylinder      & (0.01, 0.02)              & 2.232  \\
               & Thigh        & Cylinder      & (0.2, 0.05)               & 4.152  \\
               & Calf         & Cylinder      & (0.2, 0.05)               & 1.721  \\
               & Foot              & Box           & (0.28, 0.03, 0.024)       & 0.474  \\
    \midrule
    Quadruped  & Trunk                    & Box           & (0.38, 0.09, 0.11)   & 6.921  \\
               & Hip         & Cylinder      & (0.04, 0.046)             & 1.152  \\
               & Thigh       & Box           & (0.21, 0.025, 0.034)    & 1.152  \\
               & Calf        & Cylinder      & (0.12, 0.013)             & 0.154  \\
g               
               & Foot        & Sphere        & (0.022,)                  & 0.040   \\
    \midrule
    Hexapod    & Trunk                    & Box           & (0.8, 0.5, 0.1)           & 6.921  \\
               & Hip              & Sphere        & (0.05,)                   & 0.678  \\
               & Thigh            & Cylinder      & (0.22, 0.03)              & 1.152  \\
               & Calf             & Cylinder      & (0.22, 0.025)             & 0.154  \\
               & Foot             & Sphere        & (0.03,)                   & 0.040  \\
    \bottomrule
  \end{tabular}
  \label{tab:basic_units}
\end{table}

\begin{table}[h]
    \centering
    \renewcommand{\arraystretch}{1.3}
    \caption{\textbf{Motor and joint properties of the generated embodiments in \ourdataset{}}.}
    \begin{tabular}{l l r r r}
    \toprule
    Class  & Joint Name        & Joint Limits (rad) & Max. Torque (N·m) & Max. Velocity (rad/s) \\
    \midrule
    Humanoid    & Torso joint            & (-2.35, 2.35)        & 200          & 23               \\
                & Shoulder pitch joint    & (-2.87, 2.87)        & 40           & 9                \\
                & Shoulder roll joint     & (-0.34, 3.11)        & 40           & 9                \\
                & Shoulder yaw joint     & (-1.30, 4.45)       & 18           & 20               \\
                & Elbow joint            & (-1.25, 2.61)        & 18           & 20               \\
                & Hip yaw/roll joint      & (-0.43, 0.43)        & 200          & 23               \\
                & Hip pitch         & (-3.10, 2.50)        & 200          & 23               \\
                & Knee joint              & (-0.26, 2.00)        & 300          & 14               \\
                & Ankle joint            & (-0.87, 0.52)        & 40           & 9                \\
    \midrule
    Quadruped   & Hip pitch joint   & (-1.05, 1.05)        & 23.7         & 30.1             \\
                & Front thigh joint       & (-1.57, 3.49)        & 23.7         & 30.1             \\
                & Rear thigh joint       & (-0.52, 4.53)        & 23.7         & 30.1             \\
                & Knee joint        & (-2.72, -0.84)       & 45.43        & 15.7             \\
    \midrule
    Hexapod     & Hip joint         & (-1.57, 1.57)        & 100          & 30               \\
                & Thigh joint       & (-1.57, 1.57)        & 100          & 30               \\
                & Knee joint        & (-1.57, 1.57)        & 100          & 30               \\
    \bottomrule
    \end{tabular}
    \label{tab:joint_limits}
\end{table}

\begin{table}[h]
\centering
\def\arraystretch{1.1}
\caption{
\textbf{Variation parameters across geometry, topology, and kinematics in the embodiment generation algorithm.} The torso link randomization only applicable to the humanoid class. 
}
\begin{tabular}{l l l}
\toprule
Variation Type & Parameter Name & Candidate Values \\
\hline
Topology
& Number of knee joints & \{0, 1, 2, 3\} \\
\hline
Geometry
& Scaling factor for all link size  & \{0.8, 1.0, 1.2\} \\
& Scaling factor for thigh link length & \{0.4, 0.8, 1.0, 1.2, 1.6\} \\
& Scaling factor for calf link length & \{0.4, 0.8, 1.0, 1.2, 1.6\} \\
& Scaling factor for foot link size & \{1.0, 2.0\} \\
& Scaling factor for torso link size & \{0.4, 0.8, 1.0, 1.2, 1.6\} \\
\hline
Kinematics
& Scaling factor for knee joint limits & \{0.2, 0.6, 1.0\}  \\
\bottomrule
\end{tabular}

\vspace{-40pt}

\begin{flushleft}
\small
\end{flushleft}
\label{tab:gen_variations}
\end{table}


\begin{table}[h]
\centering
\def\arraystretch{1.3}
\caption{
\textbf{Nominal joint configurations for generated embodiments in \textsc{GenBot-1K}.} 
These joint angles are used to initialize robot poses, define regularization rewards, and function as offsets to the policy actions. The values are consistent across symmetric limbs.
}
\label{tab:nominal_joint_config}
\begin{tabular}{l l l}
\toprule
Class & Joint Name & Joint Angle (rad) \\
\midrule
Humanoid
& Torso & 0.0 \\
& Shoulder (Left/Right, pitch/roll/yaw) & 0.0 \\
& Elbow (Left/Right) & 0.0 \\
& Hip pitch (Left/Right) & -0.4 \\
& Hip roll/yaw (Left/Right) & 0.0 \\
& Knee (Left/Right) & 0.8 \\
& Ankle (Left/Right) & -0.4 \\
\midrule
\textbf{Quadruped}
& Hip (Front/Rear, Left/Right) & $\pm$0.1 \\
& Thigh (Front, Left/Right) & 0.8 \\
& Thigh (Rear, Left/Right) & 1.0 \\
& Knee (Front/Rear, Left/Right) & -1.5 \\
& Additional knee joints (if any) & 0.0 \\
\midrule
\textbf{Hexapod}
& Hip (Front/Middle/Rear, Left/Right) & 0.0 \\
& Thigh (Front/Middle/Rear, Left/Right) & 0.79 \\
& Knee (Front/Middle/Rear, Left/Right) & 0.79 \\
& Additional knee joints (if any) & 0.0 \\
\bottomrule
\end{tabular}
\end{table}

\clearpage

\section{Cross-Embodiment Distillation} \label{appendix:distillation}

\subsection{Expert Data Collection}
For every embodiment, we run the expert \gls{rl} policy for $600$ simulation steps using $4096$ parallel environments. This results in a total of 1,985,740,800 data samples across all training embodiments. Note that the episode length during the expert training is 1000 simulation steps (equivalent to 20 physical seconds), thus, the collected data only covers the first half of the episode. Using the full length may provide more time-correlated data, which we did not analyze due to time constraints. The final dataset needs around 5 TB of storage using the \texttt{h5py} format without additional compression.

\subsection{URMA Architecture Details}  \label{appendix:urma-architecture}
The observation space of the \gls{urma} policy is split into two parts: joint-specific observations $o_j$ and general observations $o_g$.
The joint-specific observations $o_j$ include the joint angle, joint velocity, previous action of the joint (shape: ($j(e)$, 3)).
The general observations $o_g$ include the trunk linear velocity, gravity vector, command velocities, PD gains, action scaling factor, total mass of the robot, robot dimensions, number of joints and feet size (shape: (20,)).

The description vectors $d_j$ of the joints include the relative carthesian position of the joint in the nominal configuration, joint rotation axis, joint nominal angle, maximum joint torque, maximum joint velocity, joint position limits, p-gain, d-gain and action scaling factor, robot mass and dimensions (shape: ($j(e)$, 18)).

\begin{figure}[t]
    \centering
    \includegraphics[width=\columnwidth]{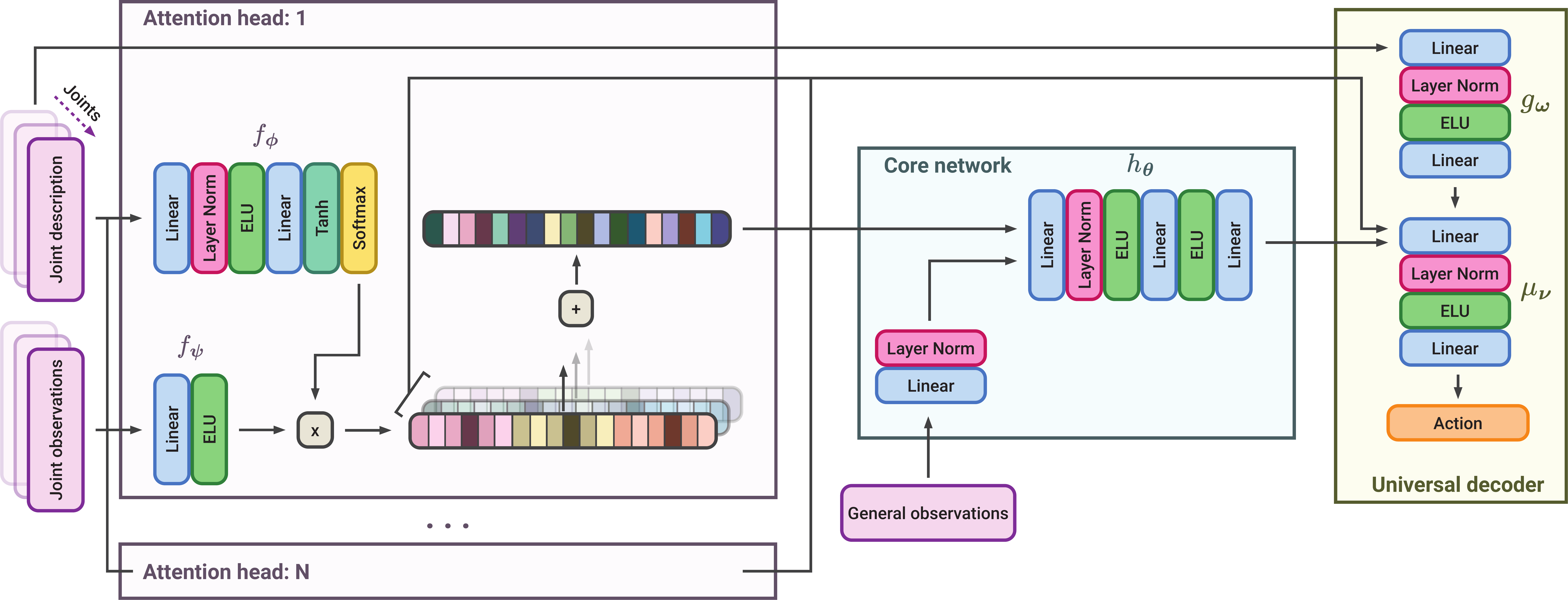}
    \caption{\textbf{\gls{urma} with multi-head attention.} 
    We extend the original \gls{urma} module~\citep{bohlinger2024onepolicy} with multiple attention heads, each aggregating information from joint observations using distinct attention distributions. This design enables the model to capture multi-modal dependencies and improves its capacity to scale across diverse embodiments.}
    \label{fig:urma_architecture}
\end{figure}

We build on the original \gls{urma} neural network architecture, as shown in Figure ~\ref{fig:urma_architecture}, from \citet{bohlinger2024onepolicy} with the following modifications:
\begin{itemize}
    \item We use multi-headed attention for the encoding of the joint observations and descriptions to increase the expressiveness of the policy. All our experiments use 3 attention heads.
    \item We remove the feet-specific attention encoder as not all robots in the real world have foot-specific sensors, like pressure sensors.
    \item We directly use the output from the action decoder $\mu_\nu$ as the action of the policy, instead of using an additional head to produce a standard deviation and sampling from a Gaussian distribution, as we train the policy with imitation learning instead of \gls{rl}.
    \item We add another encoding layer to the general observations $o_g$ to project them into a higher dimensional latent space before concatenating them with all the joint latent vectors from the attention heads.
    \item We use wider feedforward layers ($2\times$ the hidden dimensions) throughout the network. 
\end{itemize}
The resulting model has 2.1 million parameters. Overall, it is a compact network with strong inductive biases that leverage the compositional structure of robots.  

When applying the actions of the policy to the robots, we use the same PD controllers with the same nominal joint configurations and action scaling factors as in the expert training (see \appendref{sec:obs_action_space}).

\subsection{Train-Test Set Splits}
We split \ourdataset{} into a training set (80\%) and a test set (20\%) using a deterministic pseudo-random sampler with a fixed seed, ensuring full reproducibility. The same sampling procedure is applied independently to each morphology class, except for quadrupeds and hexapods, which share identical splits due to matched dataset sizes. Detailed test indices are listed in Table~\ref{tab:test_set_indices}, and summary statistics for each category are shown in Table~\ref{tab:train_test_split}.


\begin{table}[t]
\centering
\caption{\textbf{Train-test splits of \ourdataset{}.} Each index refers to one unique embodiment in each embodiment class. The training set is simply the complement of the test set and thus omitted.}
\begin{tabular}{lp{0.8\textwidth}}
\toprule
Class & Test Set \\
\midrule
Humanoid  & [0, 7, 12, 20, 31, 32, 37, 41, 46, 47, 48, 50, 51, 55, 63, 71, 72, 75, 97, 104, 111, 113, 122, 124, 128, 132, 133, 144, 149, 154, 155, 158, 161, 163, 166, 169, 170, 181, 183, 197, 204, 207, 215, 222, 226, 229, 241, 244, 248, 250, 252, 258, 260, 261, 266, 272, 276, 278, 280, 282, 286, 290, 298, 308, 312, 313, 316, 320, 327, 342] \\
Quadruped & [0, 7, 8, 20, 31, 32, 37, 41, 46, 47, 48, 50, 51, 55, 71, 72, 75, 97, 104, 111, 113, 122, 124, 128, 132, 133, 144, 149, 154, 155, 158, 161, 163, 166, 169, 170, 181, 183, 197, 204, 207, 215, 222, 226, 229, 241, 244, 248, 250, 252, 258, 260, 261, 266, 272, 278, 280, 282, 286, 290, 298, 308, 312, 313, 316, 320, 327]  \\
Hexapod   & [0, 7, 8, 20, 31, 32, 37, 41, 46, 47, 48, 50, 51, 55, 71, 72, 75, 97, 104, 111, 113, 122, 124, 128, 132, 133, 144, 149, 154, 155, 158, 161, 163, 166, 169, 170, 181, 183, 197, 204, 207, 215, 222, 226, 229, 241, 244, 248, 250, 252, 258, 260, 261, 266, 272, 278, 280, 282, 286, 290, 298, 308, 312, 313, 316, 320, 327] \\
\bottomrule
\end{tabular}
\label{tab:test_set_indices}
\end{table}

\begin{table}[t]
\centering
\caption{\textbf{Statistics of train-test splits of \ourdataset{}.} The splits have an approximately balanced distribution over different categories. }
\begin{tabular}{lccc}
\toprule
Class & Total Number & Train Set (80\%) & Test Set (20\%) \\
\midrule
Humanoid  & 348 & 278 & 70 \\
Quadruped & 332 & 265 & 67 \\
Hexapod   & 332 & 265 & 67 \\
Total     & 1012 & 808 & 204 \\
\bottomrule
\end{tabular}
\label{tab:train_test_split}
\end{table}

\subsection{Training Details}

We designed an efficient training pipeline that balances disk I/O, CPU preprocessing, GPU utilization, and RAM usage. Instead of loading every minibatch directly from disk, we first load a fixed number of data slices, each containing a small subset of steps from multiple robot embodiments, into an in-memory buffer. Each slice consists of 100 trajectories with 128 steps per trajectory. Once the buffer is filled, minibatches are sampled uniformly at random, without replacement, until every sample has been seen a fixed number of times. This strategy reduces disk access overhead, improves memory locality, and maintains sample diversity throughout training, though it may introduce local overfitting and biased gradient estimates.

Because data from different robots have varied observation and action spaces, we load them separately and use gradient accumulation to reduce bias in the gradient estimation. Specifically, gradients are accumulated across multiple minibatches before each optimizer step, helping to balance contributions across robot embodiments. While effective, this approach still suffers from local gradient bias. A more principled solution would involve zero-padding to form large, uniform batches across robots, but implementing this would require architectural and pipeline-level changes, which we did not pursue due to time constraints. In theory, this could lead to smoother optimization and potentially better final performance. 

To ensure numerical stability, we apply gradient clipping with a maximum norm of $5$. We use the AdamW optimizer~\citep{Loshchilov2019decoupled} with $\beta_1{=}0.9$, $\beta_2{=}0.999$, and a cosine-annealed weight decay schedule that decays from $3 \times 10^{-4}$ to $0$ over the course of training~\citep{Loshchilov2017sgdr}. The key hyperparameters for distillation are summarized in \tabref{tab:hyperparam_distillation}.

Our pipeline requires 128\,GB of RAM to maintain the in-memory buffer. Due to the small size of the \gls{urma} policy, training can be efficiently performed on a single GPU (e.g., NVIDIA RTX 4090 or H100). We did not observe significant gains in convergence from increasing batch size, possibly due to the structured nature and potential bottlenecks in the model architecture. Further investigation into the scaling behavior of the training dynamics is left for future work.

\begin{table}[t]
\centering
\caption{\textbf{Hyperparameters of the distillation pipeline.}}
\begin{tabular}{lc}
\toprule
Hyperparameter & Value \\
\midrule
\# training samples per embodiment & 500 $\times$ 4096 \\ 
Validation set size & 100 $\times$ 4096 \\ 
Batch size & 64 \\
Gradient accumulation steps & 8 \\
Gradient clipping threshold & 5 \\
Data slice size & 100 $\times$ 128 \\
Max slices in buffer & 1024 \\
Buffer repeat factor & 3 \\
Optimizer & AdamW~\citep{Loshchilov2019decoupled} \\
AdamW betas & $(0.9,\ 0.999)$ \\
Weight decay schedule & $3 \times 10^{-4} \rightarrow 0$ (cosine) \\
Learning rate schedule & Cosine annealing~\citep{Loshchilov2017sgdr} \\
\# epochs & 80 \\
\bottomrule
\end{tabular}
\label{tab:hyperparam_distillation}
\end{table}

\section{Additional Results on Out-of-Domain Generalization}
\label{appendix:additional_ood}

\begin{table}[h]
\centering
\def\arraystretch{1.1}
\caption{\textbf{Mean reward by knee-range scale.} Values at $0.6$ and $0.2$ are from \ourdataset{}; values at $0.1$ and $0.001$ are from the modified version.}
\label{tab:ood_rewards_by_scale}
\begin{tabular}{lcccc}
\toprule
Class & $0.6$ & $0.2$ & $0.1$ & $0.001$ \\
\midrule
Humanoid  & $19$ & $14$ & $16$ & $4$ \\
Quadruped & $49$ & $36$ & $45$ & $26$ \\
Hexapod   & $34$ & $21$ & $28$ & $20$ \\
\bottomrule
\end{tabular}
\end{table}


We evaluate out-of-distribution (OOD) generalization by testing the \gls{urma} policy trained on \ourdataset{} against a modified test distribution with significantly reduced knee joint limits—\{0.1, 0.001\}, which lie outside the training range of \{0.6, 0.2\}. Table~\ref{tab:ood_rewards_by_scale} reports the mean episode reward across morphology classes at each joint limit scale.

Results show that moderate tightening ($0.1$) induces only mild performance degradation across all classes. In contrast, extreme tightening ($0.001$) leads to a sharp drop for the less stable humanoid class, while quadrupeds and hexapods remain more robust. These results highlight the policy’s ability to generalize to structurally OOD embodiments, albeit with limitations under severe distributional shift. While expanding the training distribution to cover a broader range of joint configurations could improve robustness, such exploration is beyond the scope of this study.

\section{Additional Details on Real-World Deployment} \label{appendix:additional_real}
\subsection{Hardware Setup}
We evaluated our distilled \gls{urma} policy zero-shot on two real-world platforms: the Unitree Go2 quadruped and the Unitree H1 humanoid. For each robot, we used its URDF to produce the embodiment description vectors $d_j$.
Before deployment, the policy was converted to the ONNX format to load it in JAX and guarantee maximum inference speed. The policy inference ran on a Ubuntu 22.04 laptop (Ryzen 9 CPU), interfaced to the robot over a dedicated Ethernet connection. We ran the control loop at the same 50 Hz and with the same PD gains as in simulation, and sent the target joint angles to the robot's internal controller. We limited the commanded x-y-yaw velocity to 0.8 m/s for the Go2 and 0.5 m/s for the H1, to ensure the robot's stability and safety during the experiments.

\subsection{Implementing Joint Limit Variations}
To probe robustness under kinematic constraints, we impose an artificial knee‐joint range limited to 20, 40 or 60 \% of its nominal range. In simulation, one can enforce such limits by directly clamping joint angles within the physics engine; in hardware, however, neither the robot’s encoders nor its embedded PD controller can be modified.
Consequently, we introduce a software‐level joint‐limit layer into the control loop in order to restrict the target joint angle for affected knee joints to the new limits. At each control step, the policy’s commanded knee angle is constrained to the prescribed ±20, 40 or 60 \% bounds.
Instead, we implemented a software-based solution that restricts the target joint angle for affected knee joints to the new limits.
To counteract any excursions driven by external disturbances, we implement an active rejection mechanism: whenever the measured knee angle violates the software limits, we (1) project the commanded target onto the nearest permissible bound and (2) elevate the proportional and derivative gains to $K_p = 60$ and $K_d = 1$, respectively, until the joint re-enters the safe region. This procedure enforces a soft joint‐limit constraint exclusively in software—without altering hardware or contravening physical laws—while delivering high-gain corrective action against environmental perturbations.

\section{Additional Latent Space Analysis}
\label{appendix:further_latent_analysis}

\begin{figure}[t]
    \centering
    \includegraphics[width=0.49\columnwidth]{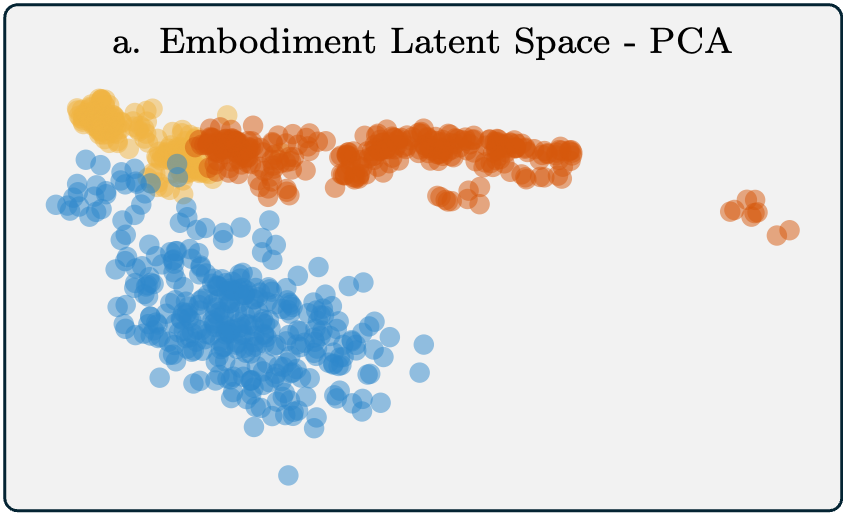}
    \includegraphics[width=0.49\columnwidth]{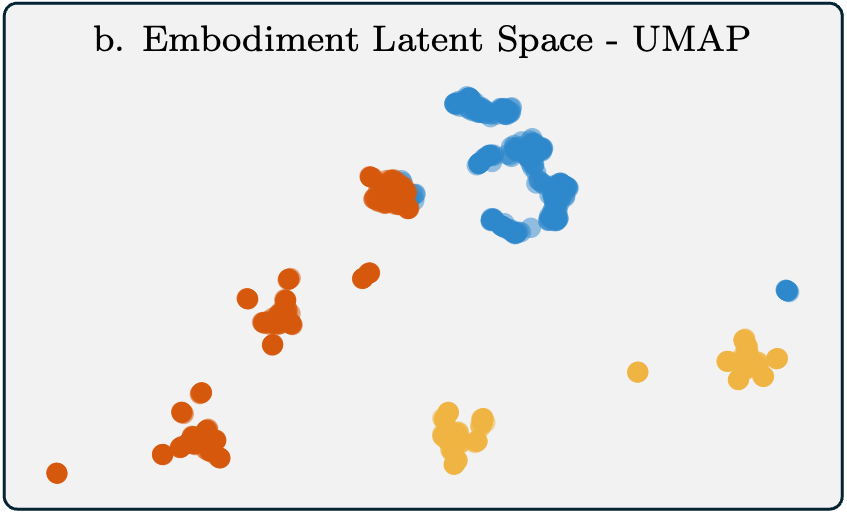}
    \hfill
    \includegraphics[width=0.51\columnwidth]{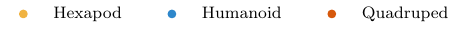}
    \caption{
        \textbf{Additional visualizations of the learned embodiment embeddings.}
        \gls{pca} (a.) and \gls{umap} (b.) of the embodiment latent space (\ie, every point represents one robot, aggregated from all of its joint description vectors).
    }
    \label{fig:further_clustering}
\end{figure}

In addition to the \gls{tsne} analysis, we also apply \gls{pca} \citep{Jolliffe2012Principal} and \gls{umap} \citep{McInnes2018UMAP} on the action latent vectors $\bar{z}_\text{action}$ in Figure~\ref{fig:further_clustering}.
Both \gls{pca} and \gls{umap} projections reveal clear grouping according to the morphology class, with humanoid, quadruped, and hexapod embeddings forming distinct clusters.
Compared to the \gls{tsne} analysis, clusters about the topological, geometric, and kinematic variations are less pronounced and appear to be more cramped.

\begin{figure}[t]
    \centering
    \includegraphics[width=\columnwidth]{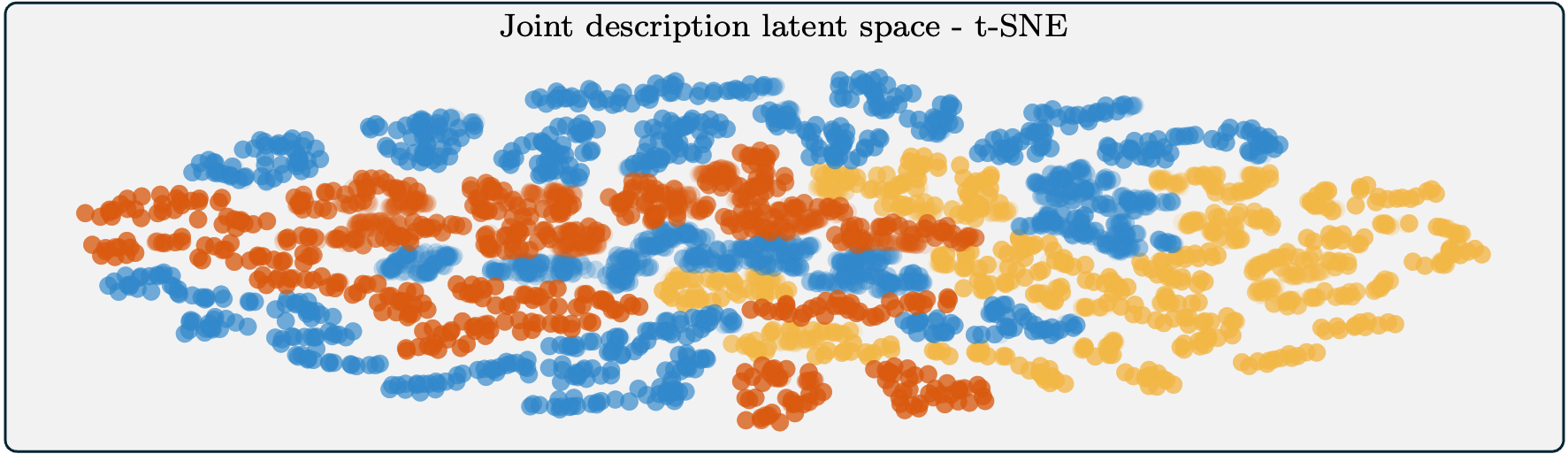}
    \hfill
    \includegraphics[width=0.51\columnwidth]{figures/legend_pca_latent.pdf}
    \caption{
        \textbf{Additional visualizations of the learned joint description embeddings.}
        \gls{tsne} visualization of the joint description latent space of all joints from all embodiments in the \ourdataset dataset (\ie, every point represents one joint of a robot).
    }
    \label{fig:joint_desc_clustering}
\end{figure}

Furthermore, we show in Figure~\ref{fig:joint_desc_clustering} the \gls{tsne} analysis of the learned joint description latent space $f_\phi(d_j)$ for all joints from all embodiments in the \ourdataset dataset.
Although the three morphologies still define the rough structure of this latent space, the learned embeddings for the joint descriptions seem to be much more entangled across the three morphology classes.

\clearpage

\vspace{-20pt}

\end{document}